\documentclass[letterpaper, 10 pt, conference]{ieeeconf}  

\IEEEoverridecommandlockouts                              

\title{\LARGE \bf
Rendering Multi-Human and Multi-Object with 3D Gaussian Splatting
}

\author{
    Weiquan Wang$^{1}$, \ 
    Jun Xiao$^{1}$, \ 
    Feifei Shao$^{1}$, \ 
    Yi Yang$^{1}$, \ 
    Yueting Zhuang$^{1}$, \ 
    Long Chen$^{2}$* \ \\
    \vspace{-0.8cm}
\thanks{This work was supported by Key R\&D Program of Zhejiang (2025C01128), the National Natural Science Foundation of China Young Scholar Fund Category B (62522216), Young Scholar Fund Category C (62402408), the National Natural Science Foundation of China (62441617, 62506333), Zhejiang Provincial Natural Science Foundation of China (No. LD25F020001), Fundamental Research Funds for the Central Universities (226-2025-00057), the Hong Kong SAR RGC General Research Fund (16219025), and Early Career Scheme (26208924), the China Postdoctoral Science Foundation (2025M781525), Postdoctoral Fellowship Program of CPSF (GZC20251077) and Zhejiang Province Postdoctoral Research Excellence Funding Project (ZJ2025065). 
(Corresponding author: Long Chen)}
\thanks{$^{1}$Weiquan Wang, Jun Xiao, Feifei Shao, Yi Yang, and Yueting Zhuang are with the State Key Lab of CAD\&CG, College of Computer
Science, Zhejiang University, Hangzhou 310027, China (e-mail:
wqwangcs@zju.edu.cn; junx@cs.zju.edu.cn; sff@zju.edu.cn; yangyics@zju.edu.cn; yzhuang@zju.edu.cn).}
\thanks{$^{2}$Long Chen is with the Department of Computer Science and Engineering,
The Hong Kong University of Science and Technology, Clear Water Bay, Hong Kong (e-mail: longchen@ust.hk).}
}

\usepackage{graphicx}
\usepackage{amsmath}
\usepackage{amssymb}
\usepackage{booktabs}
\usepackage{xcolor} 
\usepackage{colortbl}
\usepackage{multirow}

\newcommand{\eg}{\emph{e.g.},~}

\begin{document}

\maketitle
\thispagestyle{empty}
\pagestyle{empty}

\begin{abstract}

Reconstructing dynamic scenes with multiple interacting humans and objects from sparse-view inputs is a critical yet challenging task, essential for creating high-fidelity digital twins for robotics and VR/AR. This problem, which we term Multi-Human Multi-Object (MHMO) rendering, presents two significant obstacles: achieving view-consistent representations for individual instances under severe mutual occlusion, and explicitly modeling the complex and combinatorial dependencies that arise from their interactions. To overcome these challenges, we propose MM-GS, a novel hierarchical framework built upon 3D Gaussian Splatting. Our method first employs a \textbf{\emph{Per-Instance Multi-View Fusion}} module to establish a robust and consistent representation for each instance by aggregating visual information across all available views. Subsequently, a \textbf{\emph{Scene-Level Instance Interaction}} module operates on a global scene graph to reason about relationships between all participants, refining their attributes to capture subtle interaction effects. Extensive experiments on challenging datasets demonstrate that our method significantly outperforms strong baselines, producing state-of-the-art results with high-fidelity details and plausible inter-instance contacts.

\end{abstract}

\section{INTRODUCTION}

\label{sec:intro}

The development of intelligent robotic systems --- capable of operating autonomously and safely in human-centric environments --- is a central goal in robotics research~\cite{wang2024multimodal, chen2025human}.
A cornerstone for these systems is the ability to perform safe and socially-aware navigation, a task that demands a deep, fine-grained understanding of the surrounding world and the complex interactions within it~\cite{bonci2021human, lam2010human, althaus2004navigation}.
In applications ranging from assistive robotics to human-robot collaboration~\cite{cherubini2008multimode, xiao2015assistive, scheggi2016cooperative}, navigation requires that a robot must not only avoid static obstacles but also comprehend and predict the dynamics of multiple humans interacting with objects.
Creating high-fidelity, dynamic digital twins of real-world scenes serves as a critical foundation for this capability, enabling robust path planning, human-intent prediction, and advances in sim-to-real transfer~\cite{sanchez2021path, ju2022transferring, kedia2024interact}.
A critical bottleneck, however, remains in faithfully capturing and rendering the underlying 3D scene, particularly in complex, dynamic environments where multiple humans and objects interact simultaneously. This challenge gives rise to the challenging task of \textbf{Multi-Human Multi-Object (MHMO) Rendering}: the high-fidelity reconstruction of dynamic scenes from sparse-view inputs, as dense camera setups are often impractical for real-world applications.

\begin{figure}[t]
    \centering
    \includegraphics[width=\linewidth]{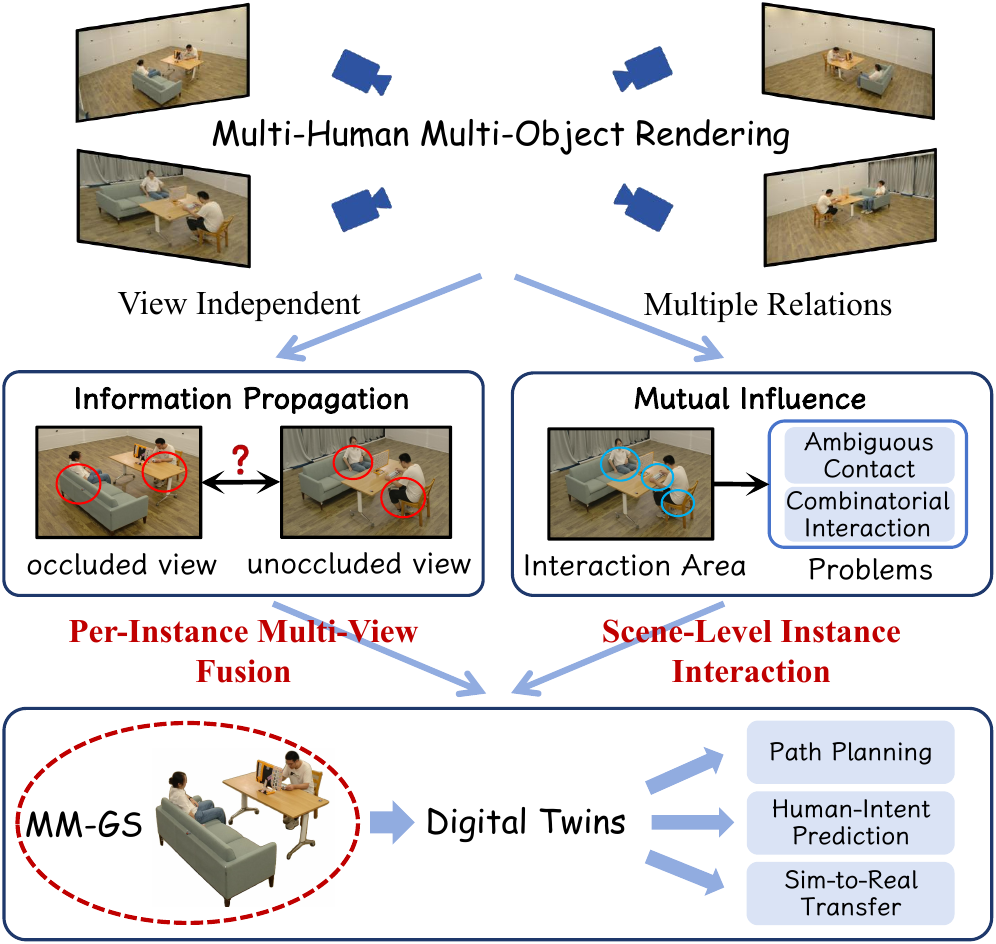}
    \caption{\textbf{Core challenges in Multi-Human Multi-Object (MHMO) rendering.} From sparse views, rendering complex interactions involves overcoming two key challenges: ensuring cross-view consistency under severe occlusion (top) and modeling the mutual influence between instances at contact regions (bottom). Our MM-GS is designed to address both.}
    \label{fig:intro_challenges}
\end{figure}


Tackling the MHMO rendering task presents two significant and coupled challenges, as illustrated in Fig.~\ref{fig:intro_challenges}. 
\textbf{The first challenge} lies in the difficulty of \emph{achieving a complete and view-consistent representation for each instance from sparse inputs}.
In a typical MHMO scene, individuals and objects frequently occlude each other, leading to severe ambiguity~\cite{zhang2024hoi}.
The optimization pipelines in prevalent methods like NeRF~\cite{mildenhall2021nerf} and 3DGS~\cite{kerbl20233d}, typically process the information from each camera view independently. 
Such a common paradigm lacks an explicit mechanism to let informative, unoccluded views guide the reconstruction of occluded ones, \eg where geometry and appearance are poorly observed or fully occluded from other angles. 
This limitation often results in artifacts or geometric inconsistencies for a single instance. For example, appearing complete from one viewpoint but blurry or fragmented from another. 
Establishing a robust and coherent per-instance representation is a critical prerequisite before their complex inter-dependencies can be modeled.

\textbf{The second challenge}, involves \emph{explicitly modeling the subtle yet crucial dependencies between interacting instances}. 
An MHMO scene is far more than a simple collection of independent entities. Its realism hinges on capturing their mutual influence~\cite{zhang2024hoi, lv2024himo}. 
These dependencies manifest primarily in resolving ambiguity at contact regions and creating subtle appearance adjustments due to proximity. 
A dedicated mechanism is required to refine the geometry and appearance of Gaussians at these contact surfaces to produce sharp, clear boundaries. 
Furthermore, close proximity between instances can introduce subtle visual effects like localized shadowing or slight color bleeding that must be captured. 
Existing works all focus on single human-object pairs~\cite{jiang2022neuralhofusion, sun2021neural, zhang2023neuraldome}, and they can only model one such relational link. 
They lack a mechanism to resolve the combinatorial complexity of MHMO scenes, where multiple humans and objects form a dense graph of potential relationships that must be reasoned about collectively to produce a coherent and realistic final rendering.

To overcome these two challenges, we propose \textbf{MM-GS}, a novel hierarchical framework built upon the efficiency and quality of 3DGS. 
Our pipeline is specifically designed to address the aforementioned problems in a coarse-to-fine manner, starting from strong human geometric priors (\eg SMPL~\cite{loper2015smpl}) and object templates. Specifically, we first align these canonical models to their target poses within each frame and then initialize a 3D Gaussian at each vertex of the resulting meshes. 
This process provides a robust initial representation for the entire scene. 
From this starting point, we tackle the first challenge of achieving view-consistency by introducing a \textbf{\emph{Per-Instance Multi-View Fusion}} module. 
Unlike standard pipelines that treat views independently, this module employs a graph attention network to create an explicit information flow between different camera views of the same instance. This allows well-observed viewpoints to inform the reconstruction of occluded or poorly-defined regions, ensuring a complete and coherent representation for each individual participant. 
Subsequently, to address the second challenge of modeling inter-instance dependencies, we introduce the \textbf{\emph{Scene-Level Instance Interaction}} module. This component constructs a global scene graph that connects all interacting instances, allowing our model to collectively reason about their dense web of relationships. By propagating information across this graph, MM-GS refines the Gaussian attributes at contact surfaces to delineate sharp boundaries and render subtle, proximity-based appearance effects, yielding a more realistic depiction of the interaction.

In summary, our main contributions are as follows:
\begin{itemize}
\item We tackle the novel and challenging problem of MHMO rendering from sparse-view inputs.
\item We propose MM-GS, a hierarchical graph-based refinement framework that models complex MHMO scenes by decoupling per-instance view-consistent representation learning from scene-level interaction modeling.
\item We design two specialized modules to explicitly address the key challenges in MHMO rendering: multi-view consistency and inter-instance dependencies.
\item Our method achieves state-of-the-art performance on complex MHMO datasets, producing realistic and coherent digital twins of human-centric environments.
\end{itemize}

\section{RELATED WORKS}
\label{sec:related}


\noindent\textbf{Free Viewpoint Rendering.}
Free Viewpoint Rendering (FVR), the task of synthesizing novel views from a set of input images~\cite{xian2021space, weng2022humannerf}, is a pivotal capability for robotics applications such as creating digital twins for simulation and policy learning~\cite{zhou2023nerf, lee2022uncertainty}. 
This field has been revolutionized by differentiable rendering, with 3DGS-based methods~\cite{kerbl20233d, zhu2025fsgs, liu2025citygaussian} offering a compelling balance of real-time speed and photorealism over prior methods like NeRF-based techniques~\cite{mildenhall2021nerf, zhou2025nerfect, li2024gp}. However, existing 3DGS-based methods lack a dedicated mechanism for the fine-grained refinement of Gaussian primitives. This makes them struggle to resolve ambiguities from severe occlusions and inter-instance dependencies, a challenge severely exacerbated in the sparse-view setting~\cite{zhang2024cor}. We are the first to address this gap, proposing a hierarchical framework that explicitly models view-consistency and inter-instance interactions to handle complex MHMO scenes from sparse views.

\noindent\textbf{Reconstruction of Dynamic and Human-Centric Scenes.}
Extending differentiable rendering techniques to dynamic environments is a significant area of research~\cite{pumarola2021d, yan2023nerf, katsumata2024compact, shih2024modeling}. A prevalent approach is to learn a deformation field that maps a static canonical representation to the observed dynamic state~\cite{park2021nerfies, liang2025gaufre, li2024deformnet, chen2025h}. For human subjects, this strategy is commonly combined with parametric body models like SMPL~\cite{loper2015smpl} to handle complex, non-rigid motion. This has enabled impressive reconstructions of single and multi-person scenes across various neural representations~\cite{weng2022humannerf, hu2023sherf, gao2022mps, qian20243dgs, liu2024humangaussian, jiang2024hifi4g}. Despite these advances, a crucial gap remains: the vast majority of existing methods treat each participant as a standalone, independent entity. They lack a mechanism to explicitly model the subtle yet crucial dependencies that arise from inter-person and person-object interactions. Our work is the first to address this gap by focusing on the high-fidelity reconstruction of complex scenes with a dense graph of inter-instance interactions.

\noindent\textbf{Modeling Human and Object Interactions.}
For a scene to be realistic, it is crucial to accurately model complex interactions between instances~\cite{collet2015high, schonberger2016structure, dou2017motion2fusion, sun2021neural, su2022robustfusion, yang2021cpf, xie2023visibility, bhatnagar2022behave}. While a line of research on human-object interaction rendering has enabled high-quality results for isolated human-object pairs, these methods are fundamentally limited in scope~\cite{jiang2022neuralhofusion, sun2021neural, jiang2023instant, wang2025physics, gavryushin2024romeo, liu2023hosnerf}. A prime example is NeuralDome~\cite{zhang2023neuraldome}, which relies on dense inputs and expensive volumetric rendering. Even recent advancements leveraging 3DGS and physical priors have focused exclusively on improving the realism of these single human-object pairs~\cite{wang2025physics}. 
A fundamental limitation of all these methods is their singular focus on a single relationship. They are not designed to handle the combinatorial complexity of realistic scenes where a dense graph of relationships exists between multiple humans and objects. In contrast, our work tackles this more general and challenging MHMO rendering problem, where a collective reasoning process over all participants is required to produce a coherent result.


\begin{figure*}[t]
    \centering
    \includegraphics[width=1.0\textwidth]{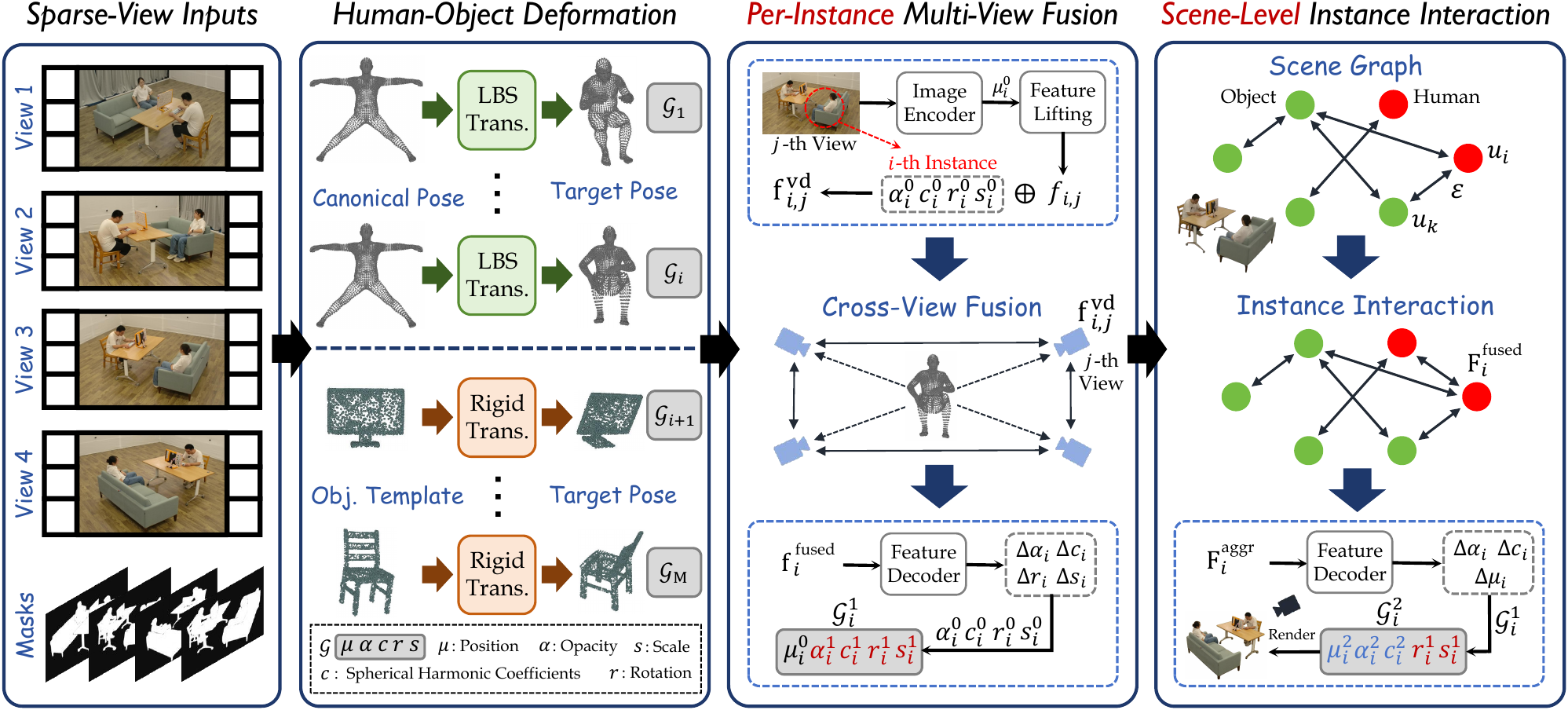}
    \caption{\textbf{Overview of MM-GS pipeline.} Our method refines initial 3D Gaussian representations through three main stages. 
    \textbf{\emph{(a) Human-Object Deformation:}} We initialize the scene by deforming canonical human and object models to their target poses and representing them as collections of 3D Gaussians. 
    \textbf{\emph{(b) Per-Instance Multi-View Fusion:}} A Cross-View Fusion network refines each instance's appearance and local geometry by aggregating visual features from all its visible viewpoints, ensuring a view-consistent representation.
    \textbf{\emph{(c) Scene-Level Instance Interaction:}} Finally, an Instance Interaction network operates on a global scene graph to model the dependencies between all participants, enabling a final refinement to capture interaction-driven effects.}
    \label{fig:overview}
\end{figure*}

\section{PRELIMINARY}
\label{sec:preliminary}

Our method is built upon the 3DGS~\cite{kerbl20233d} representation, which models a 3D scene as a collection of differentiable Gaussian primitives. This representation achieves state-of-the-art visual quality and real-time rendering speeds. Each Gaussian primitive is defined by a set of optimizable attributes $\mathcal{G} = \{\boldsymbol{\mu}, \mathbf{c}, \alpha, \mathbf{r}, \mathbf{s}\}$, which include its center $\boldsymbol{\mu} \in \mathbb{R}^3$, Spherical Harmonics (SH) coefficients for view-dependent color $\mathbf{c}$, opacity $\alpha \in \mathbb{R}$, a rotation representation $\mathbf{r}$ (e.g., a quaternion), and a scaling vector $\mathbf{s} \in \mathbb{R}^3$. The rotation and scale are used to form a covariance matrix $\boldsymbol{\Sigma}$.

To render an image, all 3D Gaussians are projected onto the 2D image plane. The color of a pixel $\mathbf{p}$ is then computed by blending the projected 2D Gaussians sorted by depth:
\begin{equation}
    \hat{C}(\mathbf{p})=\sum_{k \in \mathcal{N}} \mathbf{c}_k \sigma_k \prod_{l=1}^{k-1}(1-\sigma_l), \quad \sigma_k=\alpha_k G'_k(\mathbf{p}),
\end{equation}
where $\mathcal{N}$ is the set of sorted Gaussians overlapping pixel $\mathbf{p}$, and $G'_k$ is the projected 2D Gaussian. In this work, we manipulate the attributes of Gaussian sets belonging to different instances. We will use the notation $\mathcal{G}_i^s$ to denote the full attribute set for instance $i$ at stage $s$ of our pipeline.

\section{METHODOLOGY}
\label{sec:method}

Given $N$ sparse-view images $\{I_j\}_{j=1}^N$ of an MHMO scene with $M$ instances, along with their camera parameters and instance masks, our objective is to reconstruct and render the scene with high fidelity. Our method aims to achieve both per-instance view-consistency and realistic inter-instance interactions. To this end, our MM-GS framework, illustrated in Fig.~\ref{fig:overview}, introduces a hierarchical refinement process. First, we establish a robust initial representation by deforming canonical models to their target poses and initializing them with 3D Gaussians (Sec.~\ref{sec:deformation}). Furthermore, the \textbf{Per-Instance Multi-View Fusion} stage (Sec.~\ref{sec:view_fusion}) explicitly aggregates visual information from multiple viewpoints to ensure the completeness and consistency of each individual instance. Finally, the \textbf{Scene-Level Instance Interaction} stage (Sec.~\ref{sec:interaction}) employs a global scene graph to model the dependencies between all instances, which is crucial for resolving ambiguities at contact boundaries and capturing subtle interaction-driven appearance effects.

\subsection{Human-Object Deformation}
\label{sec:deformation}

The first stage of our pipeline establishes a robust initial geometric representation for the MHMO scene. This involves transforming all $M$ humans and objects from their canonical template spaces into their target poses, yielding the initial state (stage 0) of their Gaussian attributes, denoted as $\mathcal{G}_i^0$.

\noindent\textbf{Human Deformation.}
For each human instance $i$, we leverage the SMPL~\cite{loper2015smpl} parametric body model. We initialize 3D Gaussians on the vertices of the canonical Da-pose SMPL mesh and then deform them to the target pose via a modulated Linear Blending Skinning (LBS) process~\cite{huang2020arch, peng2021animatable, lin2022learning, liu2023hosnerf}. Specifically, the posed center $\boldsymbol{\mu}_i^0$ is computed from its canonical center $\boldsymbol{\mu}_i^c$ by applying a weighted average of each joint's transformation:
\begin{equation}
    \boldsymbol{\mu}_i^0 = \sum_{k=1}^{K} w_k (\mathbf{R}_k \boldsymbol{\mu}_i^c + \mathbf{t}_k) + \mathbf{b}, \label{eq:lbs_mean_standard}
\end{equation}
where $w_k$ is the skinning weight for the $k$-th joint and $(\mathbf{R}_k, \mathbf{t}_k)$ is the joint's transformation. 
Moreover, the covariance matrix $\boldsymbol{\Sigma}^c$ is transformed by first computing an effective blended rotation matrix, $\mathbf{r}^0$, and then applying it to the canonical covariance:
\begin{equation}
    \boldsymbol{\Sigma}^0 = \mathbf{r}^0 \boldsymbol{\Sigma}^c (\mathbf{r}^0)^{\mathrm{T}}, \quad \mathbf{r}^0 = \sum_{k=1}^{K} w_k \mathbf{R}_k. \label{eq:lbs_cov_standard}
\end{equation}

While standard LBS effectively captures the overall pose, it often struggles with fine-grained details. To address this, we employ a small MLP, $\Phi_{\text{lbs}}$, to predict a modulation vector $\mathbf{m}$ for the initial SMPL weights $w_k^{\text{SMPL}}$. The final, modulated skinning weight used in Eq.~\eqref{eq:lbs_mean_standard} is then computed as:
\begin{equation}
    w_k = \texttt{softmax}(w_k^{\text{SMPL}} + m_k).
\end{equation}
This approach yields a more detailed initial pose for each human, with the full attribute set denoted as $\mathcal{G}_i^0 = \{\boldsymbol{\mu}_{i}^0, \mathbf{c}_{i}^0, \alpha_{i}^0, \mathbf{r}_{i}^0, \mathbf{s}_{i}^0\}$, where attributes other than the center are initialized to generic values.

\noindent\textbf{Object Deformation.}
For each object instance $i$ in the MHMO scene, we assume it behaves as a rigid body. Our method directly utilizes the ground-truth pose information provided by the datasets~\cite{zhang2024hoi, liu2025core4d}. Specifically, we take the given rigid transformation, defined by a rotation matrix $\mathbf{R}_{\text{obj}}$ and a translation vector $\mathbf{T}_{\text{obj}}$, and apply it to the vertices of the object's canonical template mesh. This process accurately positions the object in the scene and defines its initial posed Gaussian centers $\boldsymbol{\mu}_i^0$. The remaining attributes are initialized to generic values to form the complete initial set $\mathcal{G}_i^0$.

\subsection{Per-Instance Multi-View Fusion}
\label{sec:view_fusion}

The deformation stage provides a robust geometric foundation $\mathcal{G}_i^0$. However, the initial attributes $\{\mathbf{c}_{i}^0, \alpha_{i}^0, \mathbf{r}_{i}^0, \mathbf{s}_{i}^0\}$ are generic and non-expressive. As established in Sec.~\ref{sec:intro}, a key challenge in MHMO rendering from sparse views is the severe ambiguity caused by inter-instance occlusions. Consequently, our primary objective in this stage is to produce a complete and view-consistent representation for each instance. To achieve this, our \textbf{\emph{Per-Instance Multi-View Fusion}} module intelligently aggregates visual information from all available viewpoints. This process operates independently on each instance $i$ and updates its state from $\mathcal{G}_i^0$ to $\mathcal{G}_i^1$.

\noindent\textbf{View-dependent Feature Construction.}
To enable context-aware fusion, we first construct a comprehensive feature representation for the instance's Gaussian set that encodes both its current 3D state and the visual evidence from each specific viewpoint. We begin by employing a 2D CNN-based image encoder $\Phi_{\text{2D}}$ to extract a multi-channel feature map $\mathbf{F}_j$ from each input image $I_j$. Then, we perform a \textit{Point-wise Feature Lifting} operation, which projects the posed Gaussian centers $\boldsymbol{\mu}_i^0$ onto the image plane of view $j$ to sample corresponding features, as formulated below:
\begin{equation}
    \mathbf{f}_{i,j}^{\text{vis}} = \mathcal{S}(\mathbf{F}_j, \Pi(\boldsymbol{\mu}_i^0, \mathbf{K}_j, \mathbf{W}_j)),
\end{equation}
where $\Pi$ is the projection function and $\mathcal{S}$ is the bilinear sampling function. Finally, this collection of visual features is concatenated with the initial optimizable attributes to form a rich and view-dependent feature representation:
\begin{equation}
    \mathbf{f}_{i,j}^{\text{vd}} = \text{Concat}(\mathbf{f}_{i,j}^{\text{vis}}, \{\mathbf{c}_{i}^0, \alpha_{i}^0, \mathbf{r}_{i}^0, \mathbf{s}_{i}^0\}).
\end{equation}

\noindent\textbf{Cross-View Fusion.}
The features $\{\mathbf{f}_{i,j}^{\text{vd}}\}_{j=1}^N$ constructed from individual views are merely isolated hypotheses. The core of our fusion process is to build a robust consensus by allowing these features to communicate and mutually refine one another. Specifically, for a Gaussian occluded in view $j$, this mechanism aggregates complementary appearance and geometric cues from visible context views $p \in \mathcal{V}_i$.
We formulate this fusion as a single operation that transforms the set of per-view features $\{\mathbf{f}_{i,j}^{\text{vd}}\}_{j=1}^{N_{\text{ctx}}}$ into a unified, view-consistent feature representation $\mathbf{f}_{i}^{\text{fused}}$:
\begin{equation}
    \mathbf{f}_{i}^{\text{fused}} = \frac{1}{N_{\text{ctx}}} \sum_{j=1}^{N_{\text{ctx}}} \left( \frac{1}{Z_{i,j}} \left( \mathbf{f}_{i,j}^{\text{vd}} + \gamma \sum_{p \in \mathcal{V}_{i}, p \neq j} \mathbf{f}_{i,p}^{\text{vd}} \right) \right),
\end{equation}
where $\gamma$ is a fusion factor, and $Z_{i,j}$ is a normalization term. $\mathbf{f}_{i}^{\text{fused}}$ serves as a robust target for the subsequent decoding.

\noindent\textbf{Fused Feature Decoding.}
The feature $\mathbf{f}_{i}^{\text{fused}}$ now encodes robust, multi-view consistent information. The final step is to decode this abstract representation into concrete updates for the Gaussian attributes. To this end, an MLP-based decoder, $\Psi_V$, maps the fused feature to the final attribute updates:
\begin{equation}
    \left( \Delta\mathbf{c}_{i}, \Delta\alpha_{i}, \Delta\mathbf{r}_{i}, \Delta\mathbf{s}_{i} \right) = \Psi_V(\mathbf{f}_{i}^{\text{fused}}).
\end{equation}
The attributes of instance $i$ are residually updated, yielding the new Gaussian set $\mathcal{G}_i^1 = \{\boldsymbol{\mu}_{i}^0, \mathbf{c}_{i}^1, \alpha_{i}^1, \mathbf{r}_{i}^1, \mathbf{s}_{i}^1\}$.



\subsection{Scene-Level Instance Interaction}
\label{sec:interaction}

The multi-view fusion stage yields a set of individually refined and view-consistent instances, denoted as $\{\mathcal{G}_i^1\}_{i=1}^M$. However, these instances are modeled in isolation and thus lack awareness of each other. As established in Sec.~\ref{sec:intro}, another fundamental challenge in MHMO rendering is explicitly modeling the subtle yet crucial dependencies between interacting entities. These dependencies are key to resolving ambiguities at contact boundaries and capturing proximity-based visual effects. To address this, our \textbf{\emph{Scene-Level Instance Interaction}} module operates on a dynamically constructed scene graph. This final stage updates the instance attributes from state $\mathcal{G}_i^1$ to their final state $\mathcal{G}_i^2$.

\noindent\textbf{Scene Graph Construction.}
To facilitate reasoning about inter-instance relationships, we first construct a scene graph $G = (\mathcal{U}, \mathcal{E})$ for each frame. The nodes $\mathcal{U} = \{u_1, ..., u_M\}$ represent the $M$ scene instances. Each node $u_i$ is represented by the instance-level feature $\mathbf{f}_{i}^{\text{fused}}$ produced by the previous fusion stage. 
The edge set $\mathcal{E}$ defines the connectivity based on spatial proximity. For each instance $i$, we compute its axis-aligned bounding box $\mathcal{B}_i = [\min(\boldsymbol{\mu}_i^0), \max(\boldsymbol{\mu}_i^0)]$, which is defined by the component-wise minimum and maximum coordinates over the set of points. An undirected edge $(u_i, u_p)$ is added to the edge set $\mathcal{E}$ if and only if the bounding boxes of the two instances intersect, resulting in a graph that connects all potentially interacting instances:
\begin{equation}
    (u_i, u_p) \in \mathcal{E} \iff \mathcal{B}_i \cap \mathcal{B}_p \neq \emptyset.
\end{equation}

\noindent\textbf{Interaction-Aware Feature Aggregation.}
With the scene graph established, we employ a graph attention network (GAT)~\cite{velivckovic2018graph} to propagate information between interacting instances. The GAT takes the set of node features $\{\mathbf{f}_{i}^{\text{fused}}\}_{i=1}^M$ as input and performs message passing along the graph edges. The network adaptively weighs the influence of neighboring instances via the attention mechanism, which learns to assign higher importance to neighbors that are physically close in contact, thereby explicitly encoding interaction dependencies. This process transforms the initial node features into a set of aggregated, interaction-aware features $\{\mathbf{f}_{i}^{\text{aggr}}\}_{i=1}^M$:
\begin{equation}
    \{\mathbf{f}_{i}^{\text{aggr}}\}_{i=1}^M = \text{GAT}(\{\mathbf{f}_{i}^{\text{fused}}\}_{i=1}^M, G).
\end{equation}
To improve efficiency, the subsequent decoding is only applied to a subset of actively interacting instances $\mathcal{U}_{\text{active}} = \{u_i \in \mathcal{U} \mid \text{deg}(u_i) >= \tau_{\text{deg}}\}$, where $\text{deg}(u_i)$ represents the degree of $i$-th node, and $\tau_{\text{deg}}$ is a pre-defined threshold.

\begin{table*}[t]
\centering
\caption{Quantitative comparison for novel view synthesis on the HOI-M$^3$ dataset. We report PSNR($\uparrow$), SSIM($\uparrow$), and LPIPS($\downarrow$) on three representative scenes. We use \colorbox{red!20}{red} and \colorbox{yellow!40}{yellow} to denote the best and second-best results.}
\label{tab:rendering_results}
\setlength{\aboverulesep}{0.5pt}
\setlength{\belowrulesep}{0.5pt}
\resizebox{\textwidth}{!}{%
\tiny
\begin{tabular}{c ccc ccc ccc}
\toprule
\multicolumn{1}{c}{\multirow{2}{*}{\textbf{Method}}} & \multicolumn{3}{c}{\textbf{Livingroom}} & \multicolumn{3}{c}{\textbf{Fitnessroom}} & \multicolumn{3}{c}{\textbf{Office}} \\
\cmidrule(lr){2-4} \cmidrule(lr){5-7} \cmidrule(lr){8-10}
& PSNR & SSIM & LPIPS & PSNR & SSIM & LPIPS & PSNR & SSIM & LPIPS \\
\midrule
NeuralHOIFVV-MM~\cite{zhang2023neuraldome} & \cellcolor{yellow!40}21.33 & \cellcolor{yellow!40}0.8704 & \cellcolor{yellow!40}0.1884 & \cellcolor{yellow!40}22.66 & \cellcolor{yellow!40}0.9148 & \cellcolor{yellow!40}0.1575 & 21.65 & \cellcolor{yellow!40}0.8876 & \cellcolor{yellow!40}0.1702 \\
GTU-MM~\cite{lee2024guess} & 20.82 & 0.8527 & 0.2045 & 21.91 & 0.9095 & 0.1683 & \cellcolor{yellow!40}21.71 & 0.8841 & 0.1734 \\
Ours & \cellcolor{red!20}22.47 & \cellcolor{red!20}0.8894 & \cellcolor{red!20}0.1722 & \cellcolor{red!20}23.24 & \cellcolor{red!20}0.9224 & \cellcolor{red!20}0.1504 & \cellcolor{red!20}22.89 & \cellcolor{red!20}0.9029 & \cellcolor{red!20}0.1633 \\
\bottomrule
\end{tabular}%
}
\end{table*}

\noindent\textbf{Gaussian Attribute Decoding.}
The aggregated feature $\mathbf{f}_{i}^{\text{aggr}}$ encapsulates the necessary contextual information to refine the Gaussian attributes in a way that reflects the interaction. Specifically, a final MLP-based decoder, $\Psi_I$, takes the interaction-aware feature $\mathbf{f}_{i}^{\text{aggr}}$ and predicts a set of residual updates for the optimizable attributes:
\begin{equation}
    \left( \Delta\boldsymbol{\mu}_{i}, \Delta\mathbf{c}_{i}, \Delta\alpha_{i} \right) = \Psi_I(\mathbf{f}_{i}^{\text{aggr}}).
\end{equation}
The final Gaussian attributes for instance $i$ are thus given by $\mathcal{G}_i^2 = \{\boldsymbol{\mu}_{i}^2, \mathbf{c}_{i}^2, \alpha_{i}^2, \mathbf{r}_{i}^1, \mathbf{s}_{i}^1\}$, where $\boldsymbol{\mu}_{i}^2 = \boldsymbol{\mu}_{i}^0 + \Delta\boldsymbol{\mu}_{i}$, $\mathbf{c}_{i}^2 = \mathbf{c}_{i}^1 + \Delta\mathbf{c}_{i}$, and $\alpha_{i}^2 = \alpha_{i}^1 + \Delta\alpha_{i}$.
Note that the position update $\Delta\boldsymbol{\mu}_{i}$ is applied to the initial posed centers $\boldsymbol{\mu}_{i}^0$ (as they were unaltered in the fusion stage), whereas the appearance updates are applied to the already-refined $\mathbf{c}_{i}^1$ and $\alpha_{i}^1$.


\subsection{Hierarchical Refinement Strategy and Loss Function}
\label{sec:strategy_loss}

Our framework's core design is a hierarchical strategy that decouples the optimization of different Gaussian attributes across stages. This staged refinement is key to resolving the complex MHMO rendering problem.

The \textbf{\emph{Per-Instance Multi-View Fusion}} stage establishes a view-consistent representation for each instance in isolation by refining its appearance ($\mathbf{c}, \alpha$) and local geometry ($\mathbf{r}, \mathbf{s}$) attributes. The Gaussian centers ($\boldsymbol{\mu}$) remain frozen during this stage for two key reasons: first, to preserve the strong geometric prior from the initial pose $\boldsymbol{\mu}^0$, and second, to provide stable spatial anchors for the feature lifting process.

The \textbf{\emph{Scene-Level Instance Interaction}} stage then addresses the relationships between instances. Here, we introduce updates to the Gaussian centers ($\boldsymbol{\mu}$) to resolve physical ambiguities like contact and penetration, which is only possible once contextual information from neighboring instances is available. We also continue to refine appearance attributes ($\mathbf{c}, \alpha$) to model effects like contact shadows. The local geometry attributes ($\mathbf{r}^1, \mathbf{s}^1$) are kept fixed, preserving the high-quality shape learned in the previous stage.

The entire framework is trained end-to-end by minimizing a composite loss, $\mathcal{L}_{\text{render}}$, between the rendered image $\hat{I}_j$ (with final attributes $\mathcal{G}^2$) and the ground-truth image $I_j$:
\begin{equation}
\begin{aligned}
    \mathcal{L}_{\text{render}} = & \lambda_{\text{L1}} ||\hat{I}_j - I_j||_1 + \lambda_{\text{SSIM}} (1 - \text{SSIM}(\hat{I}_j, I_j)) \\
    & + \lambda_{\text{LPIPS}} \mathcal{L}_{\text{LPIPS}}(\hat{I}_j, I_j),
\end{aligned}
\end{equation}
where the loss is a weighted combination of L1, SSIM~\cite{wang2004image}, and LPIPS~\cite{zhang2018unreasonable} terms.

\section{Experiments}
\label{sec:Experiments}

\subsection{Experimental Setup}
\label{sec:Experimental_Setup}


\noindent\textbf{Datasets.}
We conduct a comprehensive evaluation of our method on two challenging datasets featuring complex multi-human and object interactions.
\textbf{\emph{HOI-M$^3$}}~\cite{zhang2024hoi} is a large-scale, high-quality dataset specifically designed for capturing interactions involving multiple humans and multiple objects in various indoor scenarios. 
Our evaluation focuses on representative sequences from three diverse daily scenarios: Livingroom, Fitnessroom, and Office. For these sequences, we select four challenging sparse views (8, 12, 17, and 20) for training our model, and evaluate the novel view synthesis quality on the remaining unseen views.
\textbf{\emph{CORE4D-Real}}~\cite{liu2025core4d} is a recent dataset focusing on collaborative object rearrangement, featuring two humans interacting with a certain object. The dataset was captured using a motion capture system and four allocentric cameras. Although its scenes are slightly less complex than those in HOI-M$^3$, it provides an effective testbed for evaluating the generalization capabilities of our method. We test our method on sequences from two distinct interaction scenarios involving a bucket and a box. To demonstrate our model's robustness in these extremely sparse settings, we train our model on three randomly selected views and test on the held-out fourth view. We report the averaged results across all four possible splits.

\noindent\textbf{Implementation Details.}
Our framework is implemented in PyTorch and trained on a single NVIDIA A100 GPU using the Adam optimizer with a learning rate of $1 \times 10^{-3}$.
\textbf{\emph{Per-Instance Multi-View Fusion.}} We use a lightweight CNN for 2D feature extraction. For the cross-view propagation, the fusion factor $\gamma$ is set to $0.1$. The subsequent MLP decoder $\Psi_V$ (2 hidden layers, [128, 64], ReLU) maps the fused features and initial Gaussian attributes to residual updates and a 64-dim instance feature for the next stage. We use 4 context views for this fusion process.
\textbf{\emph{Scene-Level Instance Interaction.}} Instance interactions are modeled by a 2-layer GAT with a hidden dimension of 64, 4 attention heads, and a dropout of 0.1. The pre-defined threshold $\tau_{\text{deg}}$ is set to 1. A final MLP decoder, $\Psi_I$, maps the aggregated 64-dim GAT feature to a 7-dim modulation vector for updating the color and opacity of the Gaussians.


\begin{figure*}[t]
    \begin{center}
    \includegraphics[width=1.0\linewidth]{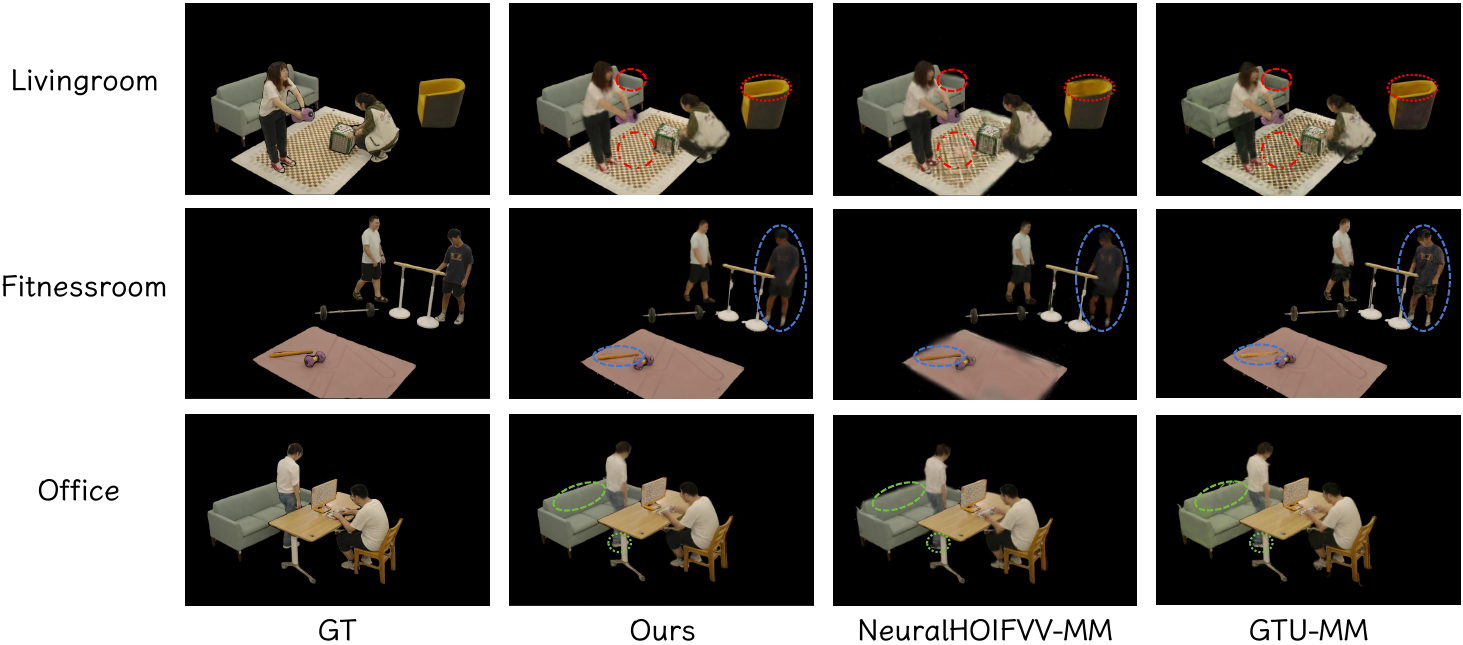}
    \end{center}
    \caption{\textbf{Qualitative comparison on the HOI-M$^3$ dataset.} We highlight specific regions with colored dashed circles to illustrate the differences. Note that our MM-GS generates significantly sharper details and more plausible contact regions. In contrast, the NeRF-based NeuralHOIFVV-MM tends to produce overly smooth or blurry results, while the 3DGS-based GTU-MM suffers from floating artifacts and geometric inconsistencies.}
    \label{fig:qualitative_results}
    \vspace{-1em}
\end{figure*}


\noindent\textbf{Baselines.}
As our work is the first to tackle the comprehensive MHMO rendering task, no direct prior methods exist for comparison. Therefore, we establish two strong baselines by extending state-of-the-art methods from related domains to our setting, ensuring a fair comparison by providing them with the same input data.
\textbf{\emph{NeuralHOIFVV-MM.}}
Our first baseline is an extension of NeuralHOIFVV~\cite{zhang2023neuraldome}, a NeRF-based method for single human-object interaction rendering. We adapt its layered representation to our multi-instance setting by assigning an independent, pose-conditioned NeRF to each human and object in the scene. All layers are then rendered compositionally. 
\textbf{\emph{GTU-MM.}}
To provide a strong comparison based on the same underlying 3D representation, we adapt GTU~\cite{lee2024guess}, a SOTA 3DGS method for multi-human reconstruction. The original method focuses solely on humans. We extend its pipeline by introducing a parallel branch for objects, where we initialize their Gaussians from posed template meshes using the provided rigid transformations. The final scene, composed of all human and object Gaussians, is then jointly optimized using the same training objectives as the original method.

\noindent\textbf{Metrics.} 
To quantitatively evaluate the quality of rendered novel view and novel pose images, we evaluate the rendering quality using standard metrics: Peak Signal-to-Noise Ratio (PSNR), Structural Similarity Index Measure (SSIM)~\cite{wang2004image}, and Learned Perceptual Image Patch Similarity (LPIPS)~\cite{zhang2018unreasonable}.



\subsection{Rendering Evaluation}

We evaluate our MM-GS against the baselines on both datasets, presenting quantitative metrics and qualitative visual comparisons.

\begin{table}[t]
\centering
\caption{Generalization performance on the CORE4D-Real dataset. Results are averaged over two interaction scenarios. Neu.-MM: NeuralHOIFVV-MM.}
\label{tab:core4d_results}
\begin{tabular}{c ccc ccc}
\toprule
\multicolumn{1}{c}{\multirow{2}{*}{\textbf{Method}}} & \multicolumn{3}{c}{\textbf{Bucket}} & \multicolumn{3}{c}{\textbf{Box}} \\
\cmidrule(lr){2-4} \cmidrule(lr){5-7}
& PSNR & SSIM & LPIPS & PSNR & SSIM & LPIPS \\
\midrule
Neu.-MM & \cellcolor{yellow!40}19.43 & \cellcolor{yellow!40}0.9347 & \cellcolor{yellow!40}0.0602 & \cellcolor{yellow!40}18.78 & 0.9218 & \cellcolor{yellow!40}0.0629 \\
GTU-MM & 19.07 & 0.9281 & 0.0644 & 18.56 & \cellcolor{yellow!40}0.9231 & 0.0637 \\
Ours & \cellcolor{red!20}20.08 & \cellcolor{red!20}0.9387 & \cellcolor{red!20}0.0527 & \cellcolor{red!20}19.22 & \cellcolor{red!20}0.9302 & \cellcolor{red!20}0.0598 \\
\bottomrule
\end{tabular}
\end{table}

\noindent\textbf{Quantitative Results.}
As shown in Table~\ref{tab:rendering_results}, our MM-GS consistently outperforms both baselines across all metrics on the challenging HOI-M$^3$ dataset. 
Notably, our approach achieves a significant improvement in PSNR and SSIM, while also attaining the lowest (best) LPIPS scores. The substantial margin over GTU-MM, which is also based on 3DGS, underscores the effectiveness of our hierarchical refinement process.
The NeRF-based NeuralHOIFVV-MM tends to produce overly smooth results, and GTU-MM, lacking our explicit fusion and interaction modules, struggles to optimize the complex scenes, leading to lower performance.
Table~\ref{tab:core4d_results} shows the generalization performance on the CORE4D-Real dataset. Even in this extremely sparse setting (training on only three views), our method continues to achieve the best results in both interaction scenarios. 

\noindent\textbf{Qualitative Results.}
The qualitative comparisons in Fig.~\ref{fig:qualitative_results} visually corroborate our quantitative superiority and highlight the specific advantages of our hierarchical design. Our method consistently generates renderings with sharp details, vibrant and consistent colors, and plausible contact regions. In contrast, NeuralHOIFVV-MM, as a NeRF-based approach, tends to produce overly smooth and blurry results, failing to capture high-frequency textures and struggling to define clear boundaries at human-object contact points. GTU-MM, while also based on 3DGS, suffers from inconsistent and mixed colors on the rendered instances. Our method avoids these issues by explicitly propagating rich visual features from the actual input views and aggregating multi-instance information, demonstrating that our proposed fusion and interaction networks are crucial for reconstructing high-fidelity MHMO scenes.



\begin{table}[t]
\centering
\caption{Ablation study of core components on the HOI-M$^3$ dataset (Livingroom).}
\label{tab:ablation_results}
\renewcommand{\arraystretch}{1.1} 
\begin{tabular}{l ccc c c}
\toprule
\textbf{Method} & PSNR$\uparrow$ & SSIM$\uparrow$ & LPIPS$\downarrow$ & FPS$\uparrow$ & Training \\
\midrule
MM-GS (Full) & \textbf{22.47} & \textbf{0.8894} & \textbf{0.1722} & 160+ & $\sim$40min \\
w/o View Fusion & 20.98 & 0.8566 & 0.1927 & 160+ & $\sim$32min \\
w/o Interaction & 21.51 & 0.8621 & 0.1893 & 160+ & $\sim$36min \\
w/o Both & 19.42 & 0.8140 & 0.2031 & 160+ & \textbf{$\sim$30min} \\
\bottomrule
\end{tabular}
\end{table}

\subsection{Ablation Study}

To validate the effectiveness of our two core components, the Per-Instance Multi-View Fusion stage and the Scene-Level Instance Interaction stage, we conduct a comprehensive ablation study on the HOI-M$^3$ dataset. We evaluate four variants of our model: (1) our full MM-GS model; (2) our model without the Scene-Level Instance Interaction stage (denoted as `w/o Interaction`); (3) our model without the Per-Instance Multi-View Fusion stage (denoted as `w/o View Fusion`); and (4) a baseline version without both modules (denoted as `w/o Both`).

\noindent\textbf{Quantitative Analysis.}
The results of our ablation study are presented in Table~\ref{tab:ablation_results}. The baseline model without either of our proposed modules performs the worst across all rendering quality metrics. Our full model, which incorporates both modules, achieves the best performance by a clear margin, confirming the complementary nature of our two-stage refinement process.
Either introducing the Per-Instance Multi-View Fusion or the Scene-Level Instance Interaction module brings significant gains over the baseline. 
Regarding computational efficiency, our full model maintains real-time rendering capabilities at over 160 FPS, demonstrating that our hierarchical design effectively balances high-fidelity interaction modeling with computational performance.

\begin{figure}[t]
    \centering
    \includegraphics[width=\linewidth]{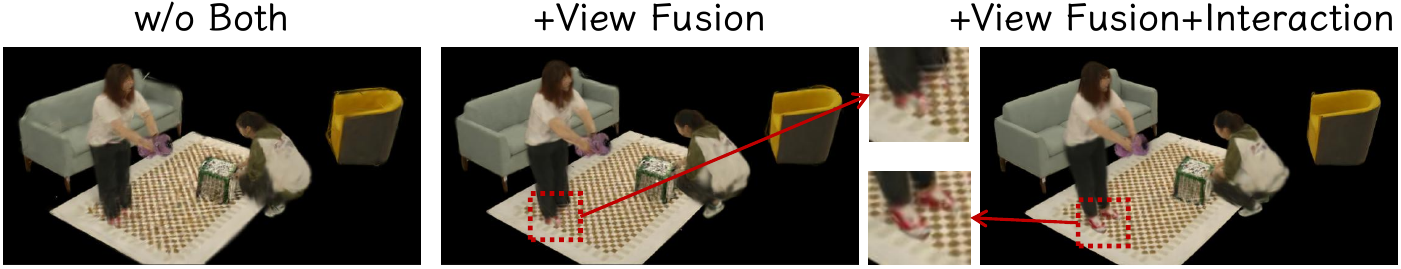}
    \caption{\textbf{Qualitative results of our ablation study.} Removing both modules (w/o Both) leads to blurry results. Adding the View Fusion module (+ View Fusion) significantly improves sharpness. Further incorporating the Interaction network (+ View Fusion + Interaction) resolves ambiguities at contact regions, resulting in cleaner boundaries.}
    \label{fig:ablation_qualitative}
\end{figure}

\noindent\textbf{Qualitative Analysis.}
The qualitative results in Fig.~\ref{fig:ablation_qualitative} provide visual evidence for the function of each module. The baseline rendering is noticeably blurry and lacks detail. After incorporating the Per-Instance Multi-View Fusion module, the sharpness and overall clarity of the rendering are significantly improved
However, ambiguities at contact points may still exist, as seen in the magnified region where the person's shoe and the carpet appear intermingled. Finally, by adding the Scene-Level Instance Interaction module, these contact boundaries become sharp and well-defined, which highlights its crucial role in reasoning about spatial relationships to produce a physically plausible and coherent scene.

\section{Conclusion}
\label{sec:conclusion}

In this paper, we introduced and tackled the novel and challenging task of rendering complex Multi-Human Multi-Object (MHMO) interactions from sparse view inputs. We presented MM-GS, a novel hierarchical framework built upon 3D Gaussian Splatting that addresses this problem through a coarse-to-fine refinement strategy. 
Extensive experiments demonstrate that our approach significantly outperforms strong baselines adapted to this new task, achieving high fidelity with plausible contacts, thereby supporting high-quality digital twin creation for robotic simulation.

\noindent\textbf{Limitations and Future Works.}
Our current scope focuses on common interaction scenarios involving articulated humans and largely rigid objects. Extending our modeling to more complex physical phenomena is a valuable avenue for future research. Moreover, our pipeline relies on available object poses and 2D masks, which pose a constraint on immediate in-the-wild robotic deployment.






\bibliographystyle{IEEEtrans}  
\bibliography{IEEEfull}

\end{document}